%% file: main.tex
\documentclass[10pt,twocolumn,letterpaper]{article}

\usepackage{cvpr}
\usepackage[table]{xcolor} 

\usepackage{booktabs} 
\usepackage{multirow} 
\usepackage{amsmath}  
\usepackage[table]{xcolor} 
\usepackage{booktabs, array, multirow}
\usepackage{booktabs} 
\usepackage{amssymb}  
\usepackage{array} 
\usepackage{adjustbox}
\usepackage{threeparttable}
\usepackage{tcolorbox}
\usepackage{xcolor}
\usepackage{verbatim}
\usepackage{listings}
\usepackage{fancyvrb}
\usepackage{nicematrix, booktabs, colortbl}
\usepackage{array}
\usepackage{ragged2e} 
\usepackage{nicematrix}
\usepackage{booktabs}
\usepackage{xcolor}
\usepackage{makecell}

\input{preamble}
\definecolor{cvprblue}{rgb}{0.21,0.49,0.74}
\usepackage[pagebackref,breaklinks,colorlinks,allcolors=cvprblue]{hyperref}

\def\paperID{4843} 
\def\confName{CVPR}
\def\confYear{2026}

\title{Symphony: A Cognitively-Inspired Multi-Agent System for Long-Video Understanding}

\author{
    Haiyang Yan\textsuperscript{1,3,*}, \quad 
    Hongyun Zhou\textsuperscript{2,*}, \quad 
    Peng Xu\textsuperscript{2}, \quad  
    Xiaoxue Feng\textsuperscript{2}, \quad 
    Mengyi Liu\textsuperscript{2,\dag} \\[1.5ex] 
    \textsuperscript{1}Institute of Automation, Chinese Academy of Sciences \quad  
    \textsuperscript{2}Kuaishou Technology \\
    \textsuperscript{3}School of Future Technology, University of Chinese Academy of Sciences  \\[1.5ex]
    \texttt{\small yanhaiyang2022@ia.ac.cn, \{zhouhongyun, xupeng09, fengxiaoxue03, liumengyi\}@kuaishou.com}
}
\date{} 

\begin{document}
\maketitle
\newcommand{\blfootnote}[1]{%
  \begingroup
  \renewcommand\thefootnote{}\footnote{#1}%
  \addtocounter{footnote}{-1}%
  \endgroup
}
\blfootnote{Work done during an internship at Kuaishou Technology. \textsuperscript{*}Equal contribution. \textsuperscript{\dag}Corresponding author: Mengyi Liu (liumengyi@kuaishou.com).}

\input{sec/0_abstract}

\input{sec/1_intro}
\input{sec/2_related}
\input{sec/3_method}
\input{sec/4_exper}

\input{sec/5_conclusion}
{
    \small
    \bibliographystyle{unsrt}
    \bibliography{main}
}
\newpage

\input{sec/6_appendix}
\end{document}

%% file: sec/0_abstract.tex
\begin{abstract}
Despite rapid developments and widespread applications of MLLM agents, they still struggle with long-form video understanding (LVU) tasks, which are characterized by high information density and extended temporal spans.
Recent research on LVU agents demonstrates that simple task decomposition and collaboration mechanisms are insufficient for long-chain reasoning tasks. Moreover, directly reducing the time context through embedding-based retrieval may lose key information of complex problems.
In this paper, we propose Symphony, a multi-agent system, to alleviate these limitations. By emulating human cognition patterns, Symphony decomposes LVU into fine-grained subtasks and incorporates a deep reasoning collaboration mechanism enhanced by reflection, effectively improving the reasoning capability. Additionally, Symphony provides a VLM-based grounding approach to analyze LVU tasks and assess the relevance of video segments, which significantly enhances the ability to locate complex problems with implicit intentions and large temporal spans.
Experimental results show that Symphony achieves state-of-the-art performance on LVBench, LongVideoBench, VideoMME, and MLVU, with a 5.0\% improvement over the prior state-of-the-art method on LVBench. Code is available at https://github.com/Haiyang0226/Symphony.
\end{abstract}

%% file: sec/1_intro.tex
\section{Introduction}
\label{sec:intro}
Long-form video understanding (LVU) is becoming increasingly important for a wide range of real-world applications, such as sports commentary, intelligent surveillance, and film analysis
\cite{cao2025videominer,doshi2020continual}. Effective LVU requires robust multimodal understanding to track entities and their relations over extended timelines and to employ advanced reasoning addressing complex queries.
\cite{chen2025scaling,zhang2025deep,wang2024lvbench,wang2025videochat}. However, as video duration increases, both information density and complexity of questions increase significantly, exacerbating the challenges in multimodal understanding and logical inference. Despite recent advances in multimodal large language models (MLLMs) \cite{maaz2023video,wang2025internvideo2,chen2024longvila,song2024moviechat}, their instruction-following and reasoning capabilities degrade with longer inputs and higher task complexity \cite{arnab2025temporal,liu2023lost,kahatapitiya2024language,hsieh2024ruler}, highlighting that accurate LVU remains a pressing and unresolved research challenge.

\begin{figure}[!t]
\centerline{\includegraphics[width=1\columnwidth]{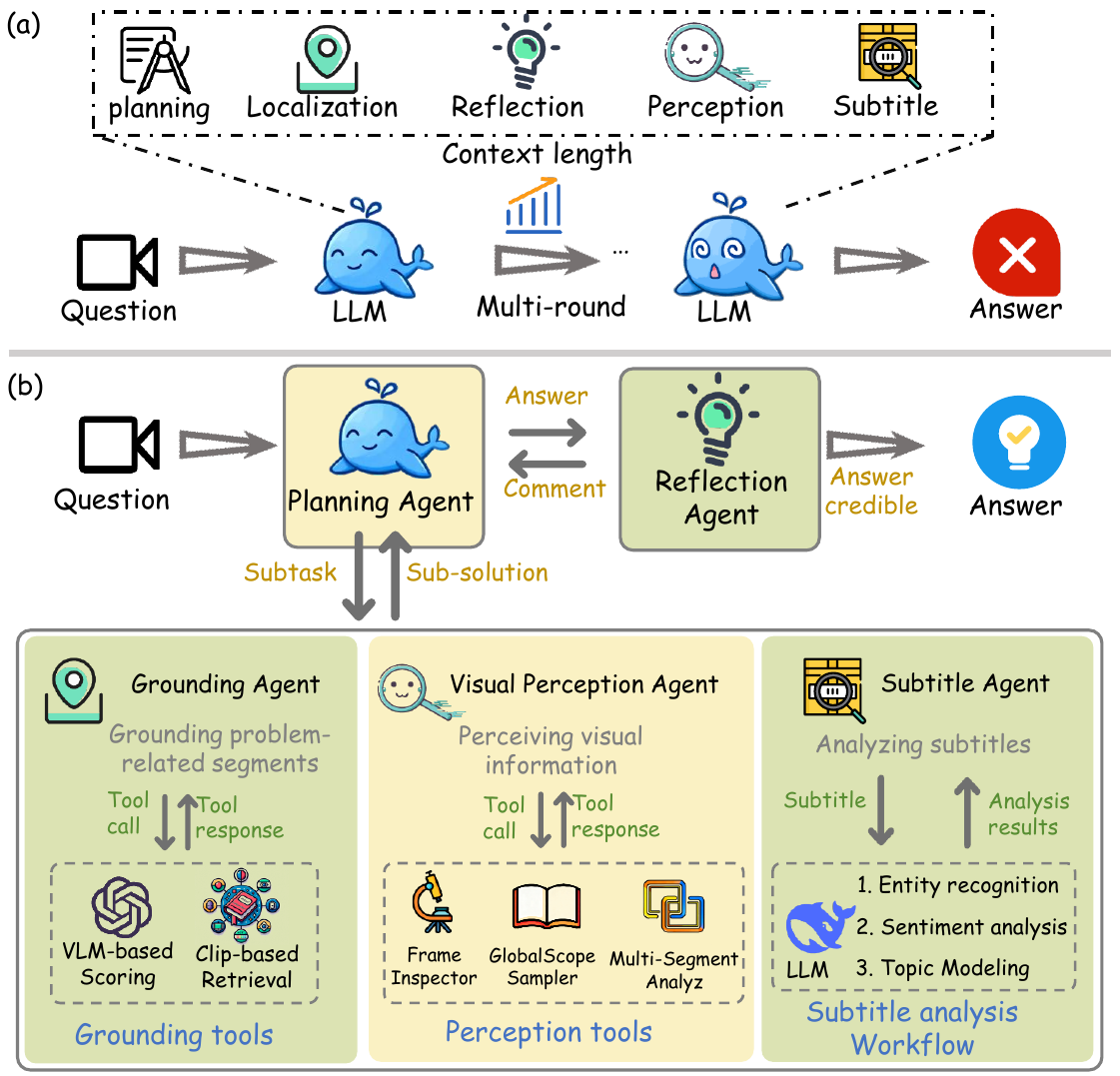}}
\caption{(a) Single-agent approaches face limitations in reasoning capacity when handling long-video understanding tasks requiring multi-step reasoning.  
(b) Our proposed multi-agent system achieves enhanced reasoning capabilities through task decomposition and collaboration along functional dimensions.}
\label{fig1}
\end{figure}

MLLM-based agents have become the predominant paradigm for LVU. Some approaches \cite{ren2025videorag,zhang2025deep,luo2024video,xu2025vrag} leverage vision-language models (VLMs) to construct video databases and employ retrieval-augmented generation (RAG) to extract relevant video segments, thereby mitigating the challenges posed by long sequences.
Nevertheless, generating effective retrieval queries from complex questions remains a significant challenge. This difficulty is further exacerbated by the prevalence of noisy and redundant content in video databases, ultimately undermining the accuracy of retrieval processes \cite{leng2024long}.
Alternatively, other works \cite{wang2024videoagent,wang2025videochat,fan2024videoagent,yuan2025videodeepresearch} leverage the reasoning capabilities of large language models (LLMs) by decomposing tasks and enabling multi-step interactions with external tools to explore the solution space. However, since all reasoning is fully delegated to the core LLM, performance degrades significantly when task complexity exceeds the model's reasoning capacity \cite{zhang2025deep,wan2025fano,stechly2024chain,shojaee2025illusion}.
Consequently, existing paradigms are hampered by limitations in grounding and reasoning, struggling to effectively tackle complex long-video tasks.

To enhance agents' capability in addressing complex problems, multi-agent systems (MAS), which harness swarm intelligence through agent collaboration, are gaining significant traction \cite{tran2025multi,fourney2024magentic,gao2025single}. VideoMultiAgent \cite{kugo2025videomultiagents} decomposes the reasoning tasks into modality-specific agents to mitigate the reasoning burden of a single agent. However, it introduces challenges in cross-modal information exchange and integration. Similarly, LvAgent \cite{chen2025lvagent} optimizes sub-agent task decomposition but relies on a linear pipeline for collaboration, which constrains the explorable task space during reasoning and fails to overcome the capability limits of single-agent methods. Consequently, while MAS holds substantial promise for LVU applications, the design of effective subtask decomposition and collaboration patterns remains a critical challenge.

To address these challenges, we propose Symphony, a centralized MAS as shown in Fig.\ref{fig1} (b). It decomposes the reasoning process of LVU tasks by emulating diverse dimensions of human cognition and orchestrates agents through a reflection-enhanced dynamic collaboration mechanism.
Specifically, the planning agent performs task decomposition and orchestrates execution by delegating sub-tasks to dedicated agents,  facilitating iterative evidence accumulation through multi-round interactions.
Subsequently, the reflection agent evaluates the reasoning chain to determine whether to output the answer or refine the reasoning process. 
This collaborative strategy substantially reduces the reasoning load on individual models with low computational overhead.
Moreover, we propose a grounding agent that leverages an LLM to decompose queries and recognize intent, coupled with a VLM to analyze video-query relevance, achieving robust and precise video grounding.
The main contributions are summarized as follows:
\begin{itemize}
\item[$\bullet$] To address the reasoning bottleneck in LVU tasks, we propose a cognitively-inspired MAS that decomposes problem-solving into specialized agents in functional dimensions. The task decomposition approach and reflection-enhanced dynamic collaboration mechanism effectively elevate the system's reasoning capabilities.

\item[$\bullet$] To improve grounding accuracy for complex questions, our proposed grounding agent leverages the reasoning capabilities of LLMs and VLMs to enable deep semantic understanding of questions and achieve more precise relevance assessment.

\item[$\bullet$] We comprehensively evaluate Symphony across four LVU datasets. On the most challenging LVBench, our approach surpasses the prior state-of-the-art method by 5.0\%. It further attains results of 77.1\%, 78.1\%, and 81.0\% on LongVideoBench, Video MME, and MLVU. 

\end{itemize}

%% file: sec/2_related.tex
\section{ Related Works }
\label{sec: Related Works}

\textbf{MLLMs for Long Video Understanding.} Recent advances in MLLMs have spurred the development of training-free strategies to address ultra-long sequence challenges. Keyframe selection methods \cite{wang2025videotree,ye2025re,he2025vsi,yang2025enhancing} employ predefined pipelines to extract question-relevant video frames, while token compression approaches \cite{wang2025adaretake,li2024videochat,liu2025video} reduce computational overhead by pruning redundant visual tokens. However, they struggle to preserve both long-term temporal semantics and fine-grained details. 
RAG-based methods \cite{luo2024video,xu2025vrag,tan2025rag,ren2025videorag,huang2025frag} 
leverage external tools to generate textual descriptions from videos, constructing retrievable databases for querying relevant snippets, yet the pre-extracted representations frequently suffer from noise and misalignment with query-specific focus.
In contrast to the single forward pass employed in prior methods, agent-based approaches offer a more comprehensive exploration of video content through planning and iterative tool invocation, thereby enabling systematic and adaptive analysis beyond superficial feature extraction.
\begin{figure*}[!t]
\centerline{\includegraphics[width=2\columnwidth]{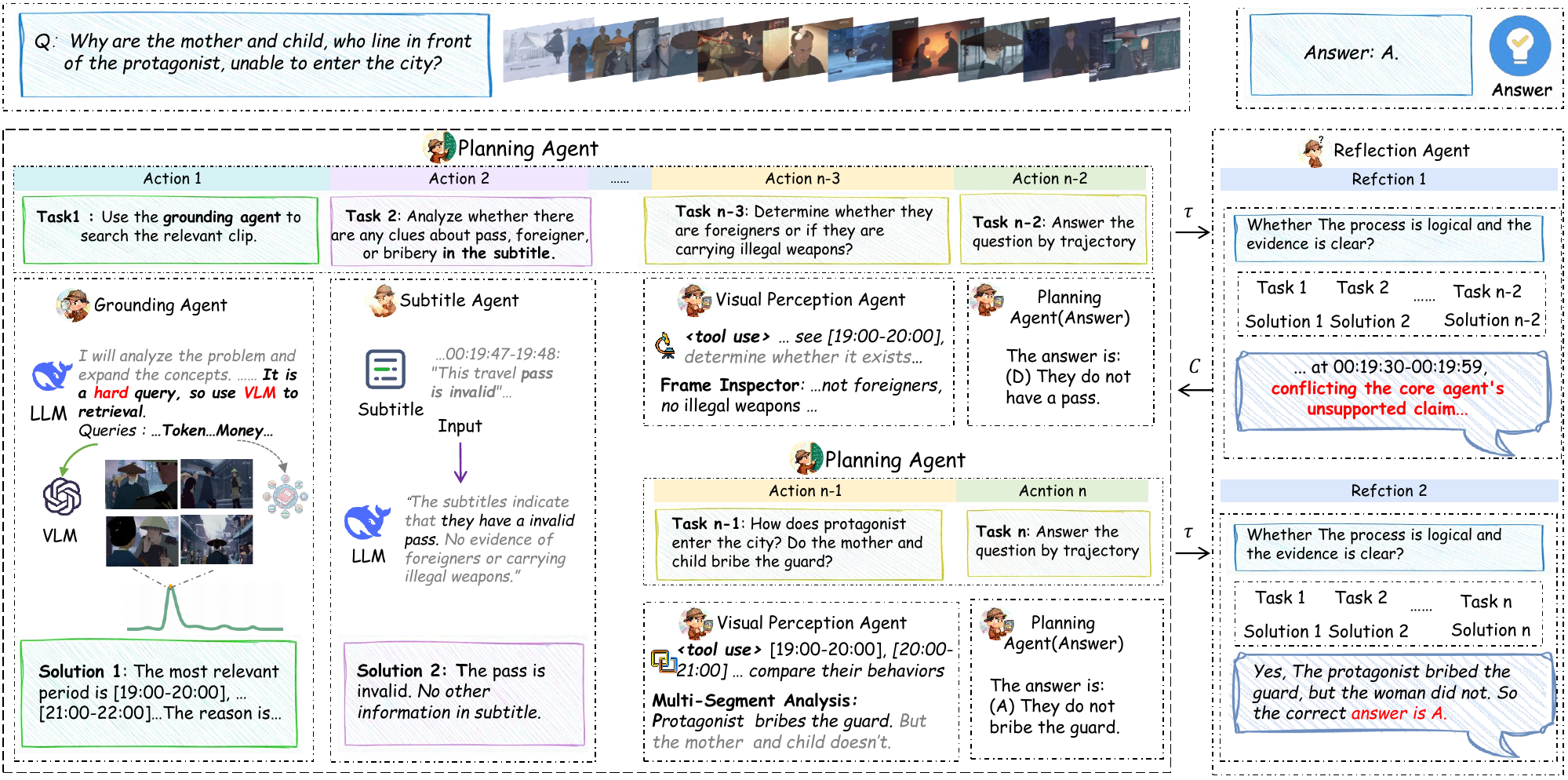}}
\caption{The reflection-enhanced dynamic reasoning framework in Symphony. The planning agent formulates a task plan and dynamically invokes other agents to execute subtasks. Upon obtaining an initial solution, the Reflection agent evaluates the reasoning chain $\tau$, producing a critique $\mathcal{C}$ that guides a subsequent round of reasoning exploration. }
\label{fig2}
\end{figure*}

\noindent \textbf{Agent-based approaches} can achieve more flexible video understanding. Specifically, VideoAgent \cite{wang2024videoagent} employs an LLM to iteratively plan and make decisions through successive rounds of CLIP-based shot retrieval and VLM perception. VideoChat-A1 \cite{wang2025videochat} introduces a chain-of-shot reasoning paradigm, employing LongCLIP \cite{zhang2024long} across successive iterations to select relevant shots and progressively subdivide them, thereby facilitating coarse-to-fine inference.
Concurrently, DVD \cite{zhang2025deep} constructs a multi-granular video database and implements tools grounded in the similarity between video captions and queries. However, these single-agent methodologies face two critical limitations: first, when task complexity surpasses the reasoning model’s capacity, the agent tends to default to simplistic actions rather than engaging in deep reasoning, rendering performance heavily reliant on costly closed-source models \cite{zhang2025deep}; second, clip-based or RAG strategies often fail to comprehensively capture query-relevant cues due to incomplete retrieval. Consequently, multi-agent systems emerge as a promising direction for achieving more accurate and robust long-video understanding.

\noindent \textbf{Multi-agent systems.} VideoMultiAgents \cite{kugo2025videomultiagents} enables collaborative multimodal reasoning by integrating specialized agents for text, visual, and graph analysis modalities. However, these agents process their respective features in isolation, failing to establish deep inter-modal interactions. LVAgent \cite{chen2025lvagent} employs a pre-selected fixed team of agents for multi-round discussions and voting, yet it adheres to a linear static workflow, which constrains the exploration of complex solution spaces and prevents dynamic, task-driven adjustments. In contrast, our method achieves deep decoupling of complex reasoning tasks along the capability dimension, significantly reducing the cognitive load for each agent. Additionally, we introduce a dedicated localization agent for precise key segment identification, providing a more accurate and robust solution for handling long videos with high spatio-temporal complexity.

%% file: sec/3_method.tex
\section{Method}
We propose Symphony, a multi-agent system composed of functionally specialized agents, as illustrated in Fig.\ref{fig1} (b). In Section 3.1, we provide an overview of the system. Section 3.2 details the collaboration mechanism among the agents, and Section 3.3 introduces our novel grounding agent.

\subsection{Overview}
Cognitive psychology traditionally decomposes human cognitive abilities into core dimensions: perception, attention, reasoning, language, and decision-making \cite{lake2017building}. Building upon this framework, we propose a capability-dimension decoupled paradigm for LVU task decomposition, implemented through a MAS. In our architecture, the planning and reflection agents jointly manage reasoning and decision-making; the grounding agent simulates the function of attention by highlighting key video segments; the subtitle agent analyzes textual subtitles to fulfill the language processing component; and the visual perception agent performs perceptual tasks.
In contrast to modality-based partitioning, which incurs high interaction costs from tight inter-module dependencies, our approach minimizes inter-agent coupling, significantly reducing the cost of information integration. This strategic allocation of cognitive load across specialized modules effectively mitigates capacity overload in monolithic architectures, enhancing accuracy and scalability in complex LVU tasks.

\begin{table}[!t]
\centering
\label{alg:liucheng}
\begin{adjustbox}{width=\columnwidth, center} 
\begin{tabular}{lp{0.48\linewidth}}
\toprule
\multicolumn{2}{l}{\textbf{Algorithm 1: Reflection-enhanced Dynamic Collaboration }} \\
\midrule
\textbf{Input}: Question $Q$, max attempts $M$ \\
\multicolumn{2}{l}{Initialize: trajectory $\tau \leftarrow \emptyset$, state $S_t \leftarrow \{\tau, Q\}$}\\
\multicolumn{2}{l}{A=\{$\mathcal{G}$ (Grounding), $\mathcal{V}$ (Visual Perception), $\mathcal{S}$ (Subtitle)\}} \\
\multicolumn{2}{l}{$m,n \leftarrow 0$} \\
\multicolumn{2}{l}{\textbf{while} $m < M$ \textbf{do}} \\
\quad \textbf{while} $n < M$ \textbf{do} \\
\quad\quad $a_t \leftarrow \text{PlanningAgent}(S_t)$ \\
\quad\quad \textbf{if} $a_t = \text{TERMINATE}$ \textbf{then break} \\
\quad\quad $o_t \leftarrow$ Execute $a_t$ using agent $\in A$ \\
\quad\quad Collect observation $o_t$ and update $\tau \leftarrow \tau \cup \{(a_t, o_t)\}$ \\
\quad\quad $n \leftarrow n + 1$ \\
\quad $\mathcal{C}, \text{Valid} \leftarrow \text{ReflectionAgent}(S_t)$ \\
\quad \textbf{if} Valid \textbf{then break} \\
\quad $S \leftarrow S \cup \{\mathcal{C}\}$, $m \leftarrow m + 1$ \\
\multicolumn{2}{l}{$A \leftarrow \text{PlanningAgent}.\text{answer}(S)$} \\
\multicolumn{2}{l}{\textbf{return} $A$} \\
\bottomrule
\end{tabular}
\end{adjustbox}
\end{table}

Specifically, the \textbf{Planning Agent} acts as the central coordinator, responsible for global task planning, multi-agent scheduling, information integration, and ultimately generating the answer. 
To efficiently and comprehensively identify question-relevant segments and potential clues within the video, the \textbf{Grounding Agent} selects either a VLM-based relevance scoring tool or a CLIP-based retrieval tool depending on the analysis of question complexity. 
The \textbf{Subtitle Agent} processes video subtitles and performs semantic analysis to enable capabilities such as entity recognition, sentiment analysis, and topic modeling.
The \textbf{Visual Perception Agent} conducts multi-dimensional visual perception by invoking three tools: frame inspector, global summary, and multi-segment analysis. Finally, the \textbf{Reflection Agent} performs a retrospective evaluation of the reasoning trajectory; upon detecting logical inconsistencies or insufficient evidence, it generates corrective suggestions to initiate a new round of refinement. The specific implementation and prompts for each agent can be found in Appendix A.

\subsection{Collaboration Mechanism}

As illustrated in Fig.\ref{fig2}, inspired by the classic Actor-Critic framework \cite{konda1999actor}, we design a reflection-enhanced dynamic reasoning framework to orchestrate the agents for solving LVU problems. The planning agent, denoted as the core policy model $\pi$, leverages the reasoning capabilities of LLM to generate sub-tasks for specialized agents, including grounding (\(\mathcal{G}\)), visual perception (\(\mathcal{V}\)), and subtitle (\(\mathcal{S}\)). Specifically, the action space for the  system is defined as 
\begin{equation}
\textbf{A} = \{\mathcal{G}, \mathcal{V}, \mathcal{S}\}.
\label{eq:forward}
\end{equation}
Motivated by recent explorations of Verifier's Law \cite{zeng2025pushing}, which posits that verifying a solution is significantly easier than generating it, we introduce a reflection agent, represented by \(\phi\) as the verification model,  to validate both the reasoning process and the final task outcomes, providing critical analysis. This mechanism expands the exploration space of MAS and improves the precision of reasoning. The collaborative mechanism of the entire system is formalized in Algorithm 1. In the following, we detail the forward reasoning process and the reflection-enhanced procedure.

In the forward reasoning stage, the planning agent evaluates and outputs the next specialized sub-task based on the state, iterating until an answer is obtained. The state $S$ comprises the initial question \(Q\) and the historical trajectory \(\tau\). For the \(t\)-th step, this process can be represented as
\begin{equation}
\begin{aligned}
a_t = \pi(S_t) \in A, \quad
S_{t}= (Q, \tau_{t-1}), \\
\tau_{t-1} = (a_1, o_1, \dots, a_{t-1}, o_{t-1}).
\end{aligned}
\label{eq:forward}
\end{equation}
Subsequently, specialized agents execute action \(a_t\) to generate output \(o_t\), update the system state \(S_t\), and iteratively carry out the forward reasoning process until sufficient evidence is accumulated to formulate an answer. During this phase, the planning agent focuses exclusively on the core logical reasoning for LVU, concentrating model capabilities to dynamically construct solution paths tailored to varying queries. This approach significantly enhances both the accuracy and efficiency of task solving compared to existing fixed workflows.

In the verification stage, the reflection agent processes \(S_T\). If the reasoning process is deemed rigorous with sufficient evidence, the procedure terminates; otherwise, the reflection agent identifies deficiencies in the reasoning chain and generates a critique formalized as $\mathcal{C} = \phi(S_T)$, followed by updates to the trajectory and state, as
\begin{equation}
\tau’_{t} = \tau’_t \cup \{\mathcal{C} \}, S'_{t}=(Q,\tau'_{t}).
\label{eq:backward}
\end{equation}
By leveraging this verification mechanism, the reflection agent re-engages the planning agent's forward reasoning process, significantly expanding the exploration space of the MAS and yielding marked performance improvements on challenging problems. 

\subsection{Grounding Agent}
\begin{figure*}[!t]
\centerline{\includegraphics[width=2\columnwidth]{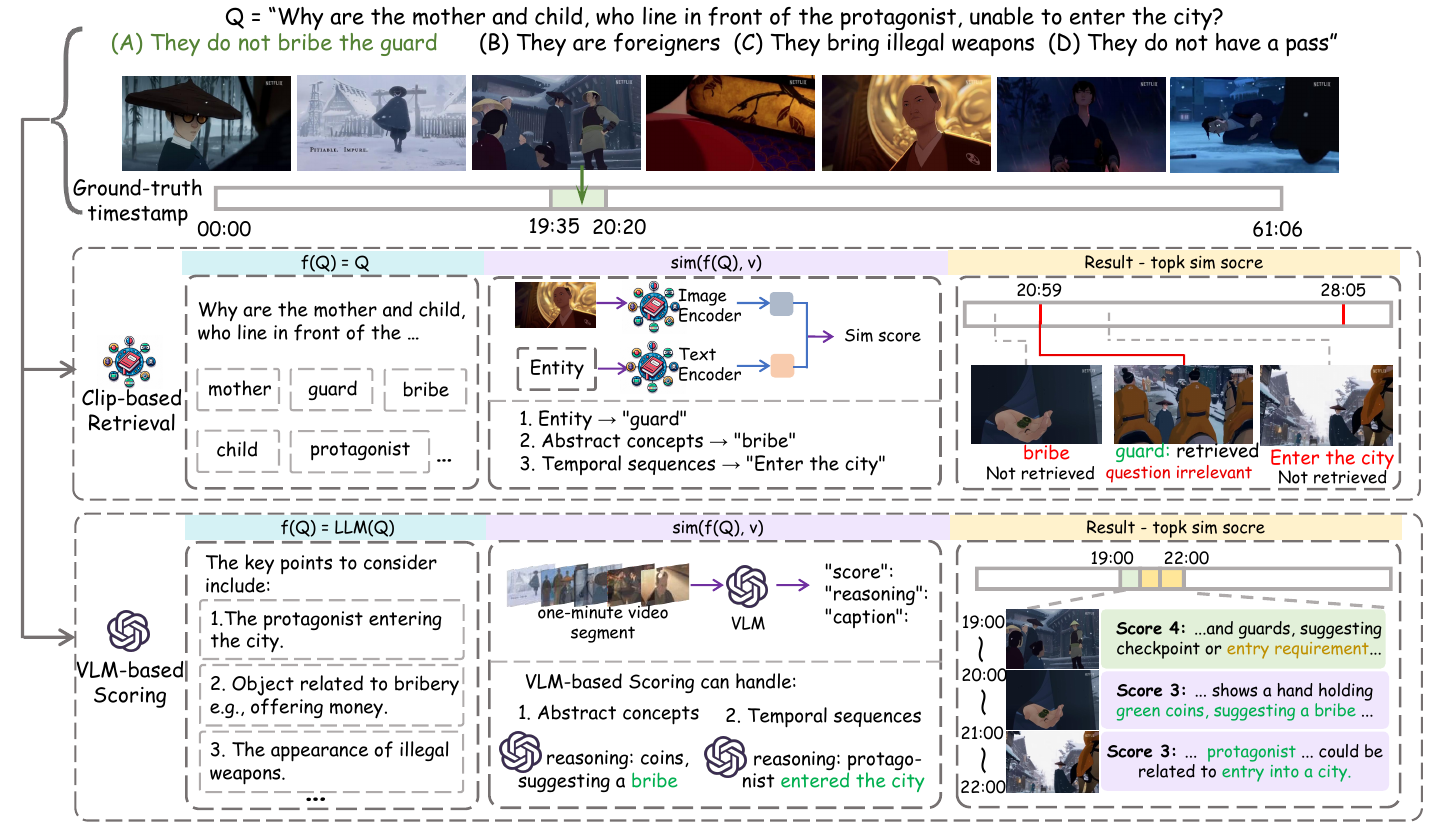}}
\caption{The CLIP-based method utilizes the original query for retrieval, thereby failing to capture abstract concepts and actions within temporal sequences. Our grounding agent analyzes the query, expands and refines the relevant concepts, and utilizes VLM to evaluate the similarity between the enhanced query and each segment, achieving more comprehensive grounding results.}
\label{fig3}
\end{figure*}

\begin{table*}[!t]
\centering
\caption{Segment relevance scoring criteria and reasoning outputs in grounding agent.}
\label{tab:scoring_criteria}
\begin{NiceTabular}{@{}c p{7.0cm} p{8.0cm}@{}}[colortbl-like]
\toprule
\textbf{Score} & \textbf{Relevance Criteria} & \textbf{Key Basis in Reasoning Process} \\
\midrule
\rowcolor{green!10}
4 & Core elements visible; sufficient for answer & Key verification steps and preliminary conclusion \\
\rowcolor{yellow!15}
3 & Partial evidence present; incomplete information & Visual cues already been identified \\
\rowcolor{orange!15}
2 & No explicit clues; indirect association through multi-hop reasoning & Approach for conceptual decomposition, semantic expansion, or associative analysis \\
\rowcolor{red!10}
1 & No observable or inferable relevance to the query & N/A \\
\bottomrule
\end{NiceTabular}
\end{table*}

Grounding aims to identify multiple video segments $S=\{s_1, s_2, \ldots, s_n\}$ from a video $V$ that are relevant to a given query, thereby eliminating interference from irrelevant content. We define a query mapping function $f$ to analyze and enhance $Q$. This process is formally defined as
\begin{equation}
S = \{ s \in V \mid \text{sim}(f(Q), s) \geq  \theta \}.
  \label{eq:also-important}
\end{equation}
where $\text{sim}(\cdot)$ denotes a function to measure the relevance between the query and video segment. 

Complex questions are prevalent in LVU tasks, thereby imposing higher demands on grounding capabilities. Specifically, this complexity is manifested in two aspects: (1) \textbf{question ambiguity}, where the question contains vague references, abstract concepts, or high-level actions without explicit entities, 
and (2) \textbf{multi-hop reasoning}, wherein the question requires chaining together multiple pieces of evidence across interrelated scenes or inferring implicit intermediate scenes. 
CLIP-based retrieval using the initial question as the query, as
\begin{equation}
f(Q)=Q, sim(f(Q),s) = \text{CLIP}(f(Q),s).
  \label{eq:also-important}
\end{equation}
Despite their effectiveness in simple questions, reliance on explicit semantic matching limits their ability to handle complex questions requiring intent recognition and logical reasoning. Therefore, we propose leveraging LLMs to generate enhanced queries, as
\begin{equation}
f(Q)\! =\! LLM(Q). 
  \label{eq:also-important}
\end{equation}
On one hand, LLMs can leverage their extensive world knowledge to disambiguate queries by instantiating ambiguous terms; on the other hand, they can utilize their reasoning capabilities to infer missing contextual information and explicitly incorporate latent logical cues.

Compared to CLIP, which relies on text-image contrastive learning, VLMs leverage more advanced cross-modal alignment strategies and exhibit enhanced reasoning capabilities. Consequently, the VLM-based similarity metric achieves a more comprehensive understanding of the semantic relationships between queries and visual content.
Therefore, we propose a VLM-based scoring tool, as
\begin{equation}
sim(f(Q),s)= \! VLM \! (f(Q),s). 
  \label{eq:also-important}
\end{equation}
Specifically, we partition the video into non-overlapping segments and perform sparse sampling within each segment. Based on the criteria shown in Table \ref{tab:scoring_criteria}, the VLM determines the relevance score and outputs the scoring rationale, with the entire process executed in parallel to reduce latency. Furthermore, the grounding agent retains the CLIP retrieval module and autonomously selects the paradigm based on query complexity.

\definecolor{best}{RGB}{176,224,230}   
\definecolor{second}{RGB}{255,239,213} 
\definecolor{header}{RGB}{70,130,180}  
\definecolor{commer}{RGB}{240,248,255} 
\definecolor{open}{RGB}{245,245,220}   
\definecolor{agent}{RGB}{255,248,220}  
\definecolor{others}{RGB}{240,248,220} 

\begin{table*}[!t]
\centering
\caption{Comparison on long video understanding benchmarks. Our method achieves SOTA performance across multiple benchmarks. The best results are \textbf{bolded}. Categories are color-coded for clarity.}
\label{long_video_benchmarks}
\renewcommand{\arraystretch}{1.15}
\setlength{\tabcolsep}{9pt}
\resizebox{0.75\textwidth}{!}{
\begin{tabular}{l c c c c}
\toprule
\rowcolor{header}
\textcolor{white}{\textbf{Methods}} & 
\textcolor{white}{LVBench} & 
\textcolor{white}{LongVideoBench (Val)} & 
\textcolor{white}{Video MME Long} & 
\textcolor{white}{MLVU} \\
\midrule

\multicolumn{5}{l}{\textit{\textbf{Commercial VLMs}}} \\
\rowcolor{commer}
Gemini-1.5-Pro \cite{team2023gemini}     & 33.1 & 64.0  & 67.4 & {-}    \\
\rowcolor{commer}
GPT-4o \cite{achiam2023gpt}               & 48.9 & 66.7  & 65.3 & 54.9 \\
\rowcolor{commer}
OpenAI o3 \cite{openai_o3_o4_mini}        & 57.1 & 67.5  & 64.7 & {-} \\

\midrule

\multicolumn{5}{l}{\textit{\textbf{Open-Source VLMs}}} \\
\rowcolor{open}
Seed 1.6 VL $^{\ast}$ \cite{guo2025seed1}                & 58.1 & 66.1  & 68.4  & 65.3 \\
\rowcolor{open}
InternVL2.5-78B\cite{wang2025internvideo2}     & 43.6 & 63.6  & 62.6  & {-}    \\
\rowcolor{open}
Qwen2.5-VL-72B \cite{bai2025qwen2}             & 47.7 & 60.7  & 63.9  & 53.8 \\

\midrule

\multicolumn{5}{l}{\textit{\textbf{Agent Based}}} \\
\rowcolor{agent}
VideoTree \cite{wang2025videotree}         & 28.8 & {-}  & 54.2 & 60.4 \\
\rowcolor{agent}
VideoAgent \cite{fan2024videoagent}        & 29.3 & {-}  & {-}  & {-} \\
\rowcolor{agent}
DVD $^{\ast}$ \cite{zhang2025deep}    & 66.8 & 67.2 & 61.5 & {-} \\
\rowcolor{agent}
VideoDeepResearch \cite{yuan2025videodeepresearch} & 55.5 & 70.6 & 76.3 & 64.5 \\
\rowcolor{agent}
VideoChatA1 \cite{wang2025videochat}       & {-} & 65.4 & 71.2 & 76.2 \\
\rowcolor{agent}
ReAgent-V \cite{zhou2025reagent}           & 41.2 & 66.4 & 72.9 & 74.2 \\

\midrule

\multicolumn{5}{l}{\textit{\textbf{Others}}} \\
\rowcolor{others}
LongVILA-7B \cite{chen2024longvila}       & {-} & 57.7  & 52.1  & 49.0 \\
\rowcolor{others}
MR. Video \cite{pang2025mr}               & 60.8 & 61.6 & 61.8  & {-} \\
\rowcolor{others}
AdaRETAKE \cite{wang2025adaretake}        & 53.3 & 67.0 & 65.0  & {-}    \\
\rowcolor{others}
VideoRAG \cite{luo2024video}              & {-} & 65.4 & 73.1 & 73.8 \\
\midrule

\textbf{Ours} &\textbf{71.8} &\textbf{77.1} &\textbf{78.1} &\textbf{81.0} \\
\bottomrule
\multicolumn{4}{l}{\small $\ast$ denotes our reproduced experimental results.}\\
\end{tabular}
}
\end{table*}

%% file: sec/4_exper.tex
\section{Experiment}

\subsection{Dataset}

We comprehensively evaluated the performance of Symphony and other state-of-the-art (SOTA) methods on four representative LVU datasets. \textbf{LVBench} \cite{wang2024lvbench} features videos with an average duration of 68 minutes, emphasizing six core capability dimensions: Temporal Grounding (TG), Summarization (Sum), Reasoning (Rea), Entity Recognition (ER), Event Understanding (EU), and Key Information Retrieval (KIR). \textbf{LongVideoBench} \cite{wu2024longvideobench} comprises 3,763 videos along with their subtitles and introduces referential reasoning tasks to evaluate fine-grained information retrieval and cross-fragment logical reasoning. 
\textbf{MLVU} \cite{zhou2024mlvu} encompasses diverse video types and is designed with nine varied tasks, including reasoning, captioning, recognition, and summarization. 
\textbf{Video-MME} \cite{fu2025video} establishes a multimodal evaluation framework spanning six broad domains, rigorously assessing spatio-temporal composite reasoning capabilities. For our experiments, we exclusively utilized the ``long" duration subset.

\subsection{Implementation Details}
Our planning and reflection agents leverage DeepSeek R1 \cite{guo2025deepseek} as the reasoning model, while the subtitle agent employs DeepSeek V3 \cite{liu2024deepseek}. The visual perception agent and grounding agent utilize Doubao Seed 1.6 VL \cite{guo2025seed1} as VLM. Input sequences are constrained to a maximum of 40 frames, with resolutions capped at 720p. For the VLM-based scoring tool, we set the duration $T = 60$ and sample 30 frames from each segment.
For MLVU and LVBench without subtitles, we used Whisper-large-v3 \cite{radford2023robust} to extract subtitles. We set the number of scheduling rounds for agents and the maximum number of tool calls within each agent to 15. For the reflection agent, the maximum number of scheduling rounds was set to 3.

\textbf{Baselines}. We comprehensively evaluated Symphony against diverse SOTA methods in LVU. The baselines include VLMs, agent-based frameworks, long-context-based LongVILA \cite{chen2024longvila}, RAG-based VideoRAG \cite{luo2024video}, and token-compression-based AdaRETAKE \cite{wang2025adaretake}. Unless specified, all results are sourced from published literature. For evaluations of Seed 1.6 VL, videos were uniformly sampled at 256 frames. For a fair comparison with DVD \cite{zhang2025deep}, we use the same reasoning model and vision model as ours.

\begin{figure*}[!t]
\centerline{\includegraphics[width=2\columnwidth]{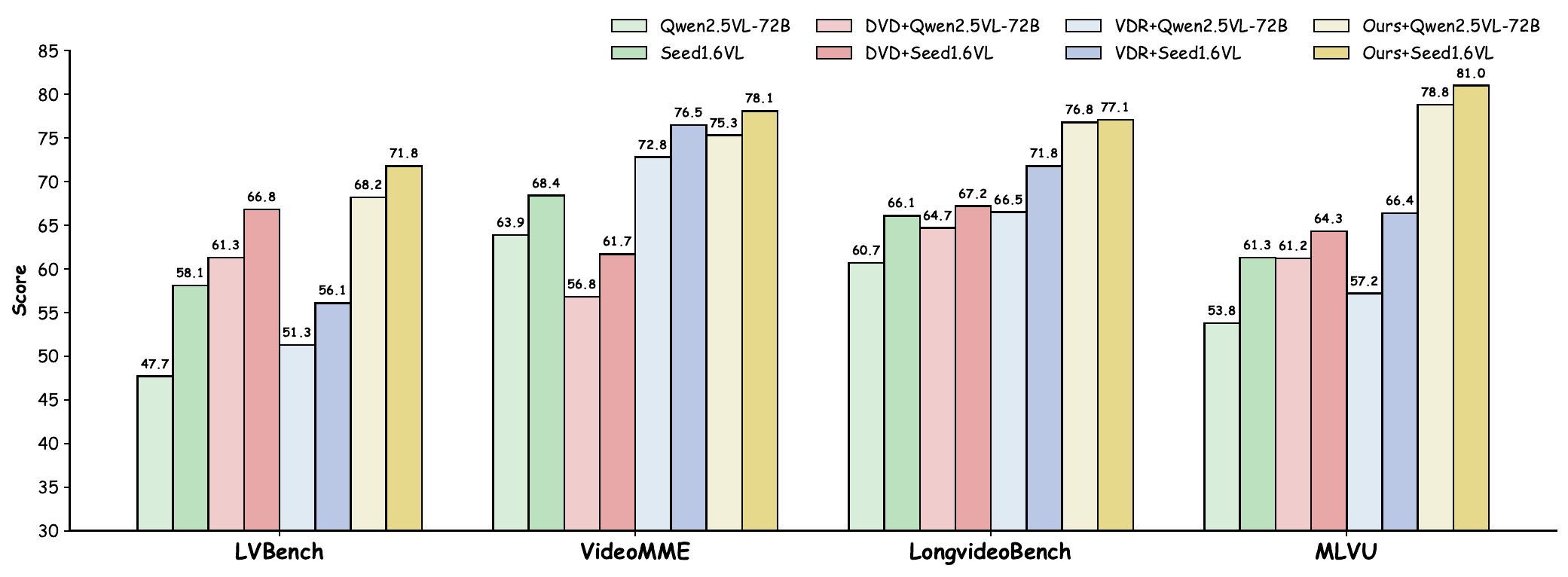}}
\caption{Experimental results of different agent-based methods.}
\label{fig_tiao}
\end{figure*}

\subsection{Main results}
The experimental results on four benchmarks are summarized in Table \ref{long_video_benchmarks}, demonstrating that Symphony consistently outperforms existing SOTA methods across all evaluations. On the most challenging LVBench, Symphony exceeds DVD by 5\%, validating the efficacy of multi-agent approaches for complex LVU tasks. Similarly, on LongVideoBench, it achieves a 6.5\% higher accuracy than VideoDeepResearch (VDR) and surpasses leading commercial VLMs. To further compare with representative agent-based methods, we evaluated DVD, VDR, and Symphony using diverse foundation models, as illustrated in Fig. \ref{fig_tiao}. These results indicate that agent-based methods generally enhance foundation models, with Symphony delivering the maximum improvement and achieving optimal outcomes independent of the foundation model. This further validates the effectiveness of the task decomposition and collaboration mechanism we proposed.

\begin{table}[!t]
\centering
\caption{Comparison of Each Capability Dimension on LVBench.}
\label{tab:lvbench}
\begin{adjustbox}{width=\columnwidth, center} 
\begin{tabular}{l c c c c c c c}
\toprule
Methods & ER & EU & KIR & TG & Rea & Sum & Overall \\
\midrule
\multicolumn{8}{l}{\textit{Commercial VLMs}} \\
Gemini-1.5-Pro \cite{team2023gemini} & 32.1 & 30.9 & 39.3 & 31.8 & 27.0 & 32.8 & 33.1 \\
GPT-4o \cite{achiam2023gpt} & 48.9 & 49.5 & 48.1 & 40.9 & 50.3 & 50.0 & 48.9 \\
OpenAI o3 \cite{openai_o3_o4_mini} & 57.6 & 56.4 & 62.9 & 46.8 & 50.8 & 67.2 & 57.1 \\
\midrule
\multicolumn{8}{l}{\textit{Open-Source VLMs}} \\
InternVL2.5-78B \cite{wang2025internvideo2} & 43.8 & 42.0 & 42.1 & 36.8 & 51.0 & 37.9 & 43.6 \\
Qwen2.5-VL-72B \cite{bai2025qwen2} & - & - & - & - & - & - & 47.7 \\
AdaRETAKE \cite{wang2025adaretake} & 53.0 & 50.7 & 62.2 & 45.5 & 54.7 & 37.9 & 53.3 \\
\midrule
\multicolumn{8}{l}{\textit{Video Agents and Others}} \\
VideoTree \cite{wang2025videotree} & 30.3 & 25.1 & 26.5 & 27.7 & 31.9 & 25.5 & 28.8 \\
VideoAgent \cite{fan2024videoagent} & 28.0 & 30.3 & 28.0 & 29.3 & 28.0 & 36.4 & 29.3 \\
MR. Video \cite{pang2025mr} & 59.8 & 57.4 & 71.4 & 58.8 & 57.7 & 50.0 & 60.8 \\
DVD \cite{zhang2025deep} & 68.2 & 65.3 & 76.0 & \textbf{70.5} & 63.1 & 61.1 & 66.8 \\
\midrule
Ours & \textbf{70.0} & \textbf{69.4} & \textbf{77.2} & 70.1 & \textbf{69.4} & \textbf{72.5} & \textbf{71.8} \\
\bottomrule
\end{tabular}
\end{adjustbox}
\end{table}

To comprehensively evaluate Symphony against other SOTA methods across different capability dimensions, we conduct a fine-grained analysis on LVBench. As shown in Table \ref{tab:lvbench}, our method achieves superior performance across multiple metrics. We attribute this superior performance to two key factors. First, the grounding agent effectively leverages the capabilities of LLM and VLM for query decomposition and spatiotemporal localization, providing rich contextual cues that enhance EU and Sum tasks. Second, the dynamic collaboration of multiple specialized agents enhances the model's reasoning capability and exploration of the solution space, resulting in an outstanding score on the Rea dimension.

\subsection{Ablation Study}
\subsubsection{Ablation Study on MAS}

To validate the effectiveness of the function-based decomposition in the proposed framework, we conduct comprehensive ablation studies comparing Symphony against its variants on the LVBench, as summarized in Table \ref{MASablation}. 

For the reflection agent, we eliminate this component and instead prompt the planning agent to perform self-reflection on its reasoning chain before generating the final answer. Our results shows that this leads to a 2.5\% performance drop. This performance gain highlights the value of an independent reflection agent, which mitigates the overconfidence common in single-agent self-correction and enhances overall reasoning reliability.

For the subtitle agent, we supply the planning agent with the complete video subtitles. This approach significantly inflates the context length, leading to a 1.4\% degradation. This highlights the role of the subtitle agent, which enables specialized analysis of subtitles while avoiding context overload in the planning agent that could impair the system's reasoning capability.

For the visual perception agent, we integrate its underlying tools into the planning agent. The results demonstrate that the specialized visual perception agent achieves a 2.2\% performance uplift by enabling more precise perception, cross-segment comparison, and nuanced analysis of visual features.

\begin{table}[!t]
\centering
\caption{Ablation Study on Each Agent in Symphony.}
\label{MASablation}
\begin{adjustbox}{width=\columnwidth, center} 

\begin{tabular}{cccc|c}
\toprule
\textbf{Planning} & \textbf{Subtitle} & \textbf{Visual Perception} & \textbf{Reflection} & \textbf{Score (\%)} \\
\midrule
\checkmark &  &  &  & 65.7 \\
\checkmark &            &  & \checkmark & 68.2 \\
\checkmark & \checkmark &            & \checkmark & 69.6 \\
\checkmark & \checkmark & \checkmark & \checkmark & \textbf{71.8} \\
\bottomrule
\end{tabular}
\end{adjustbox}
\end{table}

\subsubsection{Ablation Study on the Grounding Agent.}

\begin{table}[!t]
\centering
\caption{Ablation Study on Grounding Strategies in the Grounding Agent.}
\label{LT}
\begin{threeparttable}
\begin{adjustbox}{width=\columnwidth, center} 
\begin{tabular}{ccc|cc}
\toprule
\multicolumn{3}{c|}{Grounding Tools} & \multicolumn{2}{c}{LVBench} \\
VLM & Caption-based & Clip-based & Score ( \% )& Time(s) \\
\midrule
      -        & \checkmark  &            &  61.2 & 8.2\tnote{*} \\
      -        &             & \checkmark & 52.2  & 33.7 \\
Seed 1.6VL     &            &     &  72.1  & 68.6 \\
Qwen2.5VL-7B  &            & \checkmark &  68.6  & 37.4  \\
\midrule
Seed 1.6VL     &      & \checkmark &    71.8  & 54.8 \\
\bottomrule
\end{tabular}
\end{adjustbox}

\begin{tablenotes}
     \footnotesize
     \item[*] Excluding the time for database construction.
 \end{tablenotes}
\end{threeparttable}
\end{table}

\begin{table}[!t]
\centering
\caption{Ablation Study on Sampling Methods in the Grounding Agent.}
\label{fps}
\begin{tabular}{cc|cc}
\toprule
\multicolumn{2}{c|}{Sampling Method} & \multicolumn{2}{c}{LVBench EU Subset} \\
FPS & Clip Interval (s) & Score (\%) & Time (s)  \\
\midrule
0.5  & 30   & 70.3 & 79.0\\
0.5  & 120  & 64.7 & 50.6\\
1    & 60   & 70.9 & 74.8\\
0.25 & 60   & 62.1 & 54.8\\
\midrule
0.5  & 60 &  68.6  & 69.4 \\
\bottomrule
\end{tabular}
\end{table}

As shown in Table \ref{LT}, we conducted an ablation study on LVBench on the grounding agent. For caption-based retrieval, we utilized DVD's high-quality database generated by GPT-4.1. Although it is faster, it requires substantial time and computational resources to preconstruct the database.
Regarding the selection of VLM, we compared Seed1.6-VL with Qwen2.5VL-7B, with a concurrency count of 20. Qwen2.5VL-7B achieved inference times comparable to those of CLIP-based approaches while improving accuracy by 16.4\%, with Seed1.6-VL achieving the overall superior performance.
While relying solely on VLMs yields optimal accuracy, it results in unnecessary consumption for simple questions. In contrast, our approach yields an optimal accuracy-efficiency trade-off.

As shown in Table \ref{fps}, we ablated the parameters using the EU subset of LVBench.
Experimental results indicate that increasing FPS or reducing clip length yields only marginal accuracy improvements while significantly increasing inference latency. However, lowering FPS to 0.25 results in the omission of critical frames, causing a 6.5\% accuracy drop, whereas extending the clip interval to 120 seconds introduces irrelevant segments, failing to meet granularity requirements and leading to a 3.9\% decrease in accuracy. Our configuration achieves the optimal design.

\begin{table}[!t]
\centering
\caption{Ablation Study on Voting Strategy}
\label{voting}
\begin{adjustbox}{width=\columnwidth, center} 
\begin{tabular}{lcccc}
\toprule
Method & LVBench & LongVideo & VideoMME & MLVU \\
\midrule
LvAgent & 64.3 & 80.0 & 81.7 & \textbf{83.9} \\
Symphony & 71.8 & 77.1 & 78.1 & 81.0 \\
Symphony-Vote & \textbf{73.7} & \textbf{80.5} & \textbf{82.1} & 83.6 \\
\bottomrule
\end{tabular}
\label{tab:performance}
\end{adjustbox}
\end{table}

\subsubsection{Ablation Study on Voting Strategy}

As a general cooperative strategy, voting can effectively enhance the capabilities of agents \cite{ganaie2022ensemble}. In the field of LVU, the LvAgent \cite{chen2025lvagent} makes use of voting as part of its strategy to enhance performance. To explore the upper performance bound of our framework, we integrated a similar voting mechanism.

We simplified the three-round dynamic voting mechanism of LvAgent: three independent Symphony instances were initialized for parallel inference and majority voting.
As shown in Table \ref{voting}, we compared the performance of Symphony-Vote against the standard Symphony framework and LvAgent. The results show that the voting strategy can further enhance the proposed collaboration mechanism in Symphony, achieving a 2 to 4 percentage point gain across all benchmarks. Furthermore, Symphony-Vote outperformed LvAgent on most benchmarks.

%% file: sec/5_conclusion.tex
\section{Conclusion}
We proposed Symphony, a multi-agent system for long-form video understanding. 
Inspired by human cognitive processes, we decompose the complex LVU reasoning task into four components—planning agent, grounding agent, subtitle agent, and visual perception agent—and employ a reflection agent for the evaluation of the reasoning trajectory. This task decomposition approach, combined with a reflection-enhanced dynamic collaboration mechanism, improves the system's reasoning capability.
To improve grounding performance on complex questions, the grounding agent employs the reasoning capability of a LLM to analyze, decompose, and expand the question, and integrates a VLM-based relevance scoring tool to retrieve relevant segments.
Extensive experiments across four benchmarks demonstrate that the cognitively-inspired Symphony achieves SOTA performance. 

%% file: sec/6_appendix.tex
\appendix
\clearpage
\section*{Appendix}

\section{Details of Agents}
In this section, we present the details of the grounding agent and the visual perception agent, along with descriptions of their respective tools. Additionally, we provide the complete prompts for all agents in  \ref{pro}.

\label{sec:appendix_section}

\subsection{Grounding Agent}
To balance accuracy and efficiency, the grounding agent adaptively selects retrieval tools based on question complexity. For questions involving entities confined to a single scene (e.g., "In a room with a wall tiger and a map on the wall, there is a man wearing a white shirt. What is he doing?"), only localization within the relevant scene is required, followed by reasoning using VLMs. In such cases, the CLIP-based retrieval tool is invoked to return the top 15 most semantically similar video clips (10 seconds each segment) as grounding results. For complex questions, the VLM-based scoring tool is employed to retrieve all segments with relevance scores greater than score 1.

\subsection{Visual Perception Agent}
The visual perception agent leverages an LLM to coordinate multiple calls to a tool suite, enabling effective visual perception tasks. The toolkit comprises the following components:

\noindent \textbf{Global Summary} is designed for subtasks that require an understanding of the overall video content or thematic context. Given the video duration D, it uniformly samples 40 frames across the entire sequence and produces a compact global representation encoding high-level contextual semantics.

\noindent \textbf{Frame Inspector} is tailored for fine-grained analysis of specific temporal segments. Given a time interval $[t_s, t_e]$ as input, it performs dense frame sampling with up to 40 frames. 
The agent can employ this tool in situations requiring fine-grained, frame-level analysis, such as the inspection of temporally precise information or the examination of rapid actions.
For intervals exceeding 30 seconds in duration, an additional "cue" parameter is introduced to mitigate the risk of overlooking critical information. In addition to uniform sampling, an additional 10 frames are retrieved based on the provided cues, thereby ensuring comprehensive and robust coverage of relevant visual content.

\noindent \textbf{Multi-segment Analysis} is intended for tasks involving comparison or reasoning across non-contiguous video segments. This tool takes a list of time intervals $\left( [t_s^1, t_e^1], [t_s^2, t_e^2], \ldots, [t_s^n, t_e^n] \right)$ as input. It enables intuitive attribute comparison (e.g., changes in human appearance between temporally disjoint segments), facilitates the identification of latent informational discrepancies across segments, and supports causal analysis over time.

\subsection{Prompts for agents in Symphony}
\label{pro}
We provide the prompts for each agent in Symphony: the prompt for the planning agent is shown in Fig. \ref{PA}, that of the reflection agent in Fig. \ref{RA}, the visual perception agent in Fig. \ref{VPA}, the subtitle agent in Fig. \ref{SA}, and the grounding agent in Fig. \ref{GA}.

\section{More results}
\subsection{Case Study}
\begin{figure}[!t]
\renewcommand{\thefigure}{S\arabic{figure}} 
\centerline{\includegraphics[width=1\columnwidth]{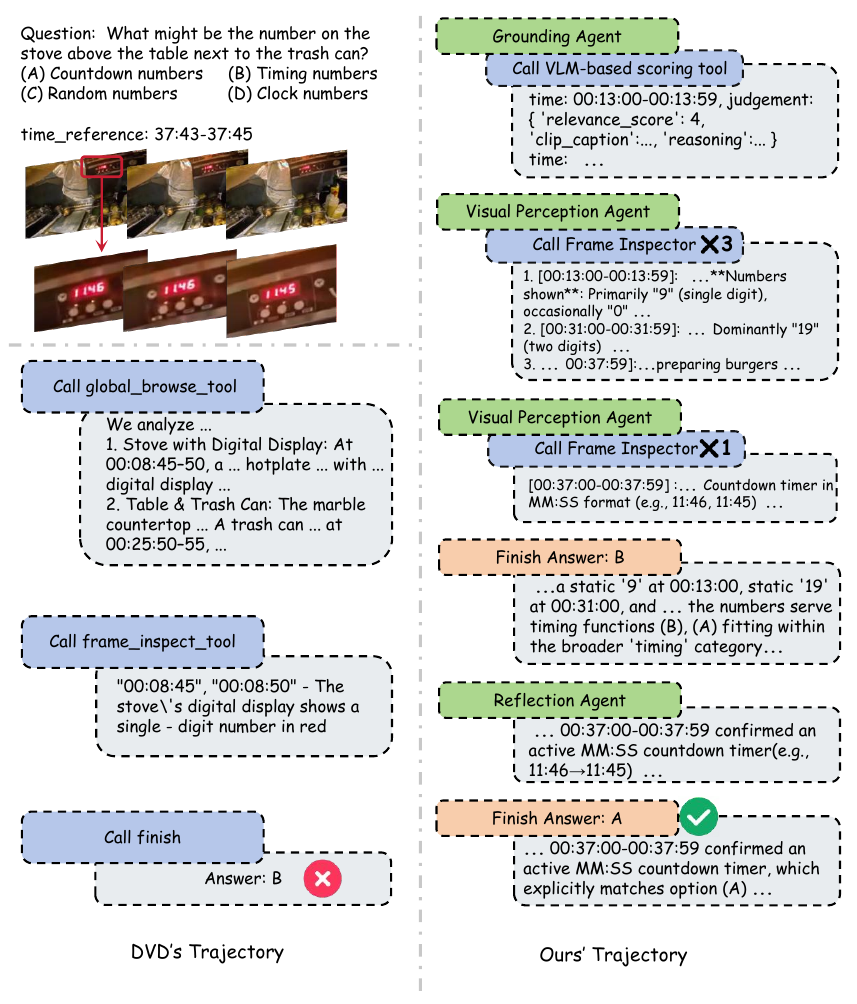}}
\caption{Analysis of the reasoning trajectories generated by our proposed Symphony and the single-agent DVD.}
\label{fig_sup}
\end{figure}

As illustrated in Fig. \ref{fig_sup}, we present the reasoning trajectories of Symphony and DVD on a complex example from LVBench. The video contains multiple scenes featuring red numerals, requiring the agent to carefully compare various visual contexts, precisely ground the red numeral located ``above the table next to the trash can," and reason over the most relevant segments to arrive at a correct answer. DVD's incomplete grounding results, combined with the model's overconfidence, lead to an incorrect response. In contrast, Symphony achieves accurate performance by leveraging its grounding agent to identify key segments with high recall, followed by iterative perception and cross-scene comparison through the visual perception agent. 
Therefore, more accurate grounding provides a solid foundation for correct answers. Meanwhile, the decomposition of agents based on capability and the collaboration mechanism foster improved reasoning abilities, ultimately enabling precise inference.

\subsection{Foundation Models}
In the comparison of existing SOTA methods, we report their published results as the strongest performance. To eliminate the influence of the foundation model, we re-evaluate existing open-source agent methods using various models. As shown in Table~\ref{tab:comparison}, our method still achieves the best performance. These results demonstrate that the performance improvement stems from the design of our agent system.

\begin{table}[t]
\centering
\caption{Performance on LVBench with different base models.}
\label{tab:comparison}
\small
\setlength{\tabcolsep}{4pt}
\begin{tabular*}{\linewidth}{@{\extracolsep{\fill}}lcccc@{}}
\toprule
Method &
\makecell{Seed\\1.6VL} &
\makecell{Qwen2.5VL\\-72B} &
\makecell{Qwen2.5VL\\-7B} &
\makecell{GPT\\4o} \\
\midrule
VideoTree   & 33.7 & 32.0 & 29.4 & 32.8\\
VideoAgent  & 37.6 & 34.6 & 30.5 & 32.7\\
VDR         & 56.1 & 52.3 & 49.4 & 50.8\\
VideoRAG    & 59.2 & 57.7 & 50.2 & 52.3\\
Ours        & \textbf{71.8} & \textbf{68.2} & \textbf{65.1} & \textbf{67.1}\\
\bottomrule
\end{tabular*}
\end{table}

\section{Cost}

We evaluate both our proposed Symphony and other agent-based methods using the LLM API provided by Alibaba Cloud. On LVBench, we evaluated the cost of DeepSeek R1, which constitutes the primary component of our method. On average, each query consumes 0.22 million tokens, amounting to \$0.124. This represents a 41.8\% reduction compared to the \$0.213 cost per query of DVD using OpenAI o3 as the reasoning model. The reduced cost is attributed to our approach leveraging more cost-effective open-source models.

\begin{figure*}[h]
\renewcommand{\thefigure}{S\arabic{figure}}
  \center
\begin{tcolorbox}[colback=gray!5!white,colframe=gray!75!black,title=Prompt for Planning Agent,top=0pt,bottom=0pt]
\begin{lstlisting}[
    basicstyle=\ttfamily\scriptsize,
    breaklines=true,  % 允许自动换行
    breakatwhitespace=true,
    frame=none,
    abovecaptionskip=5pt, 
]
You are a **Planning Agent** responsible for orchestrating the solution of complex long-video understanding tasks through systematic reasoning and dynamic collaboration within a multi-agent framework. As the central cognitive and coordination module, you are tasked with decomposing high-level queries into executable subtasks, managing information flow across specialized agents, and synthesizing evidence into accurate, well-supported conclusions.

Given a user question Q and a historical trajectory of previously executed actions and their corresponding observations, your primary responsibilities are as follows:

1. **Question Analysis**:  
   Perform a thorough semantic and logical analysis of Q. Identify its core components, implicit assumptions, temporal or causal dependencies, and domain-specific concepts. Determine whether the query pertains to visual content, events, object interactions, actions, dialogue, or higher-order reasoning (e.g., intent inference, narrative comprehension).

2. **Task Decomposition and Reasoning Strategy**:  
   Decompose the main problem into a sequence of fine-grained, logically ordered subtasks that collectively form a valid reasoning path toward answering Q. Each subtask must be actionable, contextually grounded, and designed to reduce uncertainty or fill knowledge gaps.

3. **Dynamic Planning and Contextual Decision-Making**:  
   Base each planning decision on the current state of accumulated evidence. Evaluate the sufficiency, consistency, and confidence level of existing observations. If information is incomplete, ambiguous, or insufficient to proceed with high confidence, generate the next most informative subtask that directly addresses the critical knowledge gap.

4. **Agent Orchestration**:  
   Select and delegate subtasks to specialized agent based on the required information:
   
   - **Grounding Agent**: Invoked to determine the precise timestamp(s) of a specific event, action, object appearance, or scene transition in the video. Use this agent when no explicit time range is provided in the query and temporal localization is necessary.
     
   - **Visual Perception Agent**: Utilized to analyze visual content within a specified time interval. Capable of recognizing objects, identifying actions, describing scenes, tracking spatial relationships, and answering detailed visual questions. Requires both a clear instruction and a defined time range for processing.
     
   - **Subtitle Agent**: Employed to retrieve, extract, and analyze textual subtitles associated with the video. Suitable for understanding spoken dialogue, contextual narratives, or text-based cues relevant to the question.

5. **Termination and Final Response Generation**:  
   Once the accumulated observations provide sufficient, consistent, and high-confidence evidence to fully answer Q, invoke the `finish` action. 

**Critical Rules:**
1. Your operation must be **adaptive**, **evidence-driven**, and **goal-directed**, maintaining an explicit reasoning trace throughout the process. Prioritize efficiency by minimizing redundant queries and maximizing information gain per step.
2. When conflicting information arises, the visual perception agent should be employed to compare disparate segments, identify inconsistencies, and resolve contradictions, thereby deriving a coherent and reliable conclusion.
3. Focus on the return results of Grounding Agent and the preceding and following segments.
4. The scene descriptions provided by the Grounding Agent **must** be double-checked by using the Visual Perception Agent.

Call Agents in json format:**
{{
    "reason": "Why this agent is the best choice for the task.",
    "agent": "Specific Agent name",
    "instruct": "Specific questions regarding the video"
}}

The user's question is: "{question}"
Video duration: "{duration}"

Here is the execution history:
<history>
{history_str}
</history>

\end{lstlisting}
\end{tcolorbox}
\caption{Prompt for Planning Agent.}
\label{PA}
\end{figure*}

\begin{figure*}[h]
\renewcommand{\thefigure}{S\arabic{figure}}
  \center
\begin{tcolorbox}[colback=gray!5!white,colframe=gray!75!black,title=Prompt for Reflection Agent,top=0pt,bottom=0pt]
\begin{lstlisting}[
    basicstyle=\ttfamily\scriptsize,
    breaklines=true,  % 允许自动换行
    breakatwhitespace=true,
    frame=none,
    abovecaptionskip=5pt, 
]
Please evaluate the credibility of the entire problem-solving process and the proposed answer based on the following information:
Operations performed by the core agent to solve a video understanding problem include:
{history}

The original video understanding question:
Question: {question}

The final answer proposed by the core agent:
Proposed Answer: {proposed_answer}

Evaluation Criteria:
- If the process and answer are credible and correct, set "credible"to true.
- If any errors are found, set "credible"to falseand provide a concise explanation stating what the issue is and why the proposed answer is incorrect.

Please respond strictly in the following JSON format:
{{
    "credible": boolean, // true means the answer is credible, false means it is not
    "comment": "Your concise explanation. This should be null if credible is true"
}}

Please return only the JSON object.
\end{lstlisting}
\end{tcolorbox}
\caption{Prompt for Reflection Agent.}
\label{RA}
\end{figure*}

\begin{figure*}[h]
\renewcommand{\thefigure}{S\arabic{figure}}
  \center
\begin{tcolorbox}[colback=gray!5!white,colframe=gray!75!black,title=Prompt for Subtitle Agent,top=0pt,bottom=0pt]
\begin{lstlisting}[
    basicstyle=\ttfamily\scriptsize,
    breaklines=true,  % 允许自动换行
    breakatwhitespace=true,
    frame=none,
    abovecaptionskip=5pt, 
]
You are a specialized Subtitle Analysis agent. Your task is to analyze the video subtitles based on the user's question.

Based on the following information:
The original video understanding question:  {question}
The full video subtitles for analysis: {subtitles}

Your Analysis Task:
1. Question-elevant Analysis: Extract subtitle segments directly related to the question from the original subtitles.
2. Entity and Sentiment Identification: Use the subtitle information to identify key entities mentioned and their associated sentiment.
3. General Content Summary: Provide a brief, high-level summary of the overall topic covered in the subtitle content.

Please respond strictly in the following JSON format:
{{
  "relevant_subtitle_info": "A multi-line string containing the most relevant subtitle segments. Format each entry as:\n[HH:MM:SS - HH:MM:SS]: Actual subtitle text.\nFor example:\n[00:15:32 - 00:15:35]: ...\n[00:18:05 - 00:18:09]: ...",
  "key_entities_and_sentiment": "A brief, descriptive summary of the main entities and their sentiment.",
  "overall_topic": "A one-sentence summary of the main topic discussed in the video, based only on the subtitles."
}}

Please return only the JSON object.
\end{lstlisting}
\end{tcolorbox}
\caption{Prompt for Subtitle Agent.}
\label{SA}
\end{figure*}

\begin{figure*}[h]
\renewcommand{\thefigure}{S\arabic{figure}} 
  \center
\begin{tcolorbox}[colback=gray!5!white,colframe=gray!75!black,title=Prompt for Visual Perception Agent,top=0pt,bottom=0pt]
\begin{lstlisting}[
    basicstyle=\ttfamily\scriptsize,
    breaklines=true,  % 允许自动换行
    breakatwhitespace=true,
    frame=none,
    abovecaptionskip=5pt, 
]
You are an agent responsible for video content perception. You will receive an Instruct from an upstream agent.
**Instruct:**
<Instruct>
Instruct_PLACEHOLDER
</Instruct>

**Task:**
Follow the Instruct and use tools to analyze video content to obtain key information.

**Tool Usage Guidelines:**  
*   **Video Multimodal Content Viewing:**
- To retrieve detailed information, call the frame_inspector with the time range [HH:MM:SS, HH:MM:SS]. Ensure the time range is longer than 10 seconds and less than 60 seconds. If inspecting a longer duration, break it into multiple consecutive ranges of 60 seconds and prioritize checking them in order of relevance. The end time should not exceed the total duration of the video.
- If you want to obtain a rough overview / background of a long period of time (entire video, or time range more than 3 minutes), use the global_summary_tool.
- If the (question and options) includes multi scenes, call the multi_segment_analysis_tool with a list of time range to get the answer.

**Invocation Rules:**
1.  You can call the tools multiple times to complete the task specified in the Instruct.
2.  Call only one tool at a time.
3.  Do not include unnecessary line breaks in the tool parameters.
4.  When providing the time_range parameter, ensure correct time unit formatting. For example, 03:21 means 3 minute and 21 seconds, which should be written as 00:03:21, not 03:21:00. Pay special attention to this.

**Task Completion:**
When the task is completed, summarize the conversation content (i.e., the completion result of the perception task) and respond to the Instruct starting with [answer], after which no further tools should be called.
\end{lstlisting}
\end{tcolorbox}
\caption{Prompt for Visual Perception Agent.}
\label{VPA}
\end{figure*}

\begin{figure*}[h]
\renewcommand{\thefigure}{S\arabic{figure}}
  \center
\begin{tcolorbox}[colback=gray!5!white,colframe=gray!75!black,title=Prompt for Grounding Agent,top=0pt,bottom=0pt]
\begin{lstlisting}[
    basicstyle=\ttfamily\scriptsize,
    breaklines=true,  % 允许自动换行
    breakatwhitespace=true,
    frame=none,
    abovecaptionskip=5pt, 
]
You are an agent responsible for localizing temporal segments in a video that are relevant to a given question. First, analyze the question, then select the appropriate tool based on its type, and generate enhanced queries.

## Question Information
- Question: QUESTION_PLACEHOLDER
- Video Duration: VIDEO_LENGTH

## Question Analysis Process
When presented with a query, you must conduct a thorough analysis to determine its complexity level, focusing on two critical dimensions: question ambiguity and multi-hop reasoning requirements.
1. Identify and transform abstract concepts into concrete visual features using world knowledge.
2. Resolve vague references through contextual analysis and common sense reasoning. 
3. Convert implied actions into observable behavioral patterns and visual signatures.

## Tool Descriptions
### retrieve_tool
- Description: Retrieves the most relevant time points from a video based on a textual cue.
- Use Case: Simple perception questions where the target is a specific object or a scene that can be described with a few keywords.
- Parameters:
- - cue: A short descriptive text.
- - frame_path: Path to the video frames.
- Returns: A list of timestamps.

### vlm_scoring_tool
- Description: Designed for more complex questions that require a deeper understanding of the video content, such as identifying actions, events, or scenarios.
- Use Case: Complex questions requiring scenario understanding.
- Parameters:
- - question: The question to be answered.
- - scoring_instruction: A detailed description, based on the Question Analysis, of what to identify in the video.
- - frame_path: Path to the video frames.
- Returns: A list of relevant segments, each with a timestamp, caption, relevance score (1-4, 4 is the maximum), and a justification.

## Tool Selection based on Question Type
- Type 1: The question does not involve any action, is a simple perception question, and contains detailed scene/character descriptions. The character references are clear, and there is no ambiguity in the question, call retrieve_tool for scene localization.
- Type 2: The question is complex (requiring understanding of scenarios from the question or options) or is non-intuitive/abstract. Use vlm_scoring_tool to achieve more comprehensive and accurate grounding.

### finish
- Description: Returns the localization result.
- Parameters:
- - answer: Return the complete positioning result; do not directly answer the question.

\end{lstlisting}
\end{tcolorbox}
\caption{Prompt for Grounding Agent.}
\label{GA}
\end{figure*}

\begin{figure*}[h]
\renewcommand{\thefigure}{S\arabic{figure}}
  \center
\begin{tcolorbox}[colback=gray!5!white,colframe=gray!75!black,title=Prompt for vlm\_scoring\_tool,top=0pt,bottom=0pt]
\begin{lstlisting}[
    basicstyle=\ttfamily\scriptsize,
    breaklines=true,  % 允许自动换行
    breakatwhitespace=true,
    frame=none,
    abovecaptionskip=5pt, 
]
You are given a sequence of video frames sampled from a 1-minute video clip and a question. 
Your task is to:
1. Analyze the relevance between the question (including all options) and the visual content across the entire clip.
2. Output a global relevance score and description.

Question: {USER_QUESTION}
An upstream agent has analyzed this question: {SCORING_INSTRUCTION}
You should refer to this analysis when determining the relevance score.

Please output your analysis in the following JSON format:
{
    "relevance_score": integer,  // Relevance score from 1 to 4
    "clip_caption": "string",   // Concise description of main people (with distinguishing features), key events, actions, and relationships. Focus on elements related to the question.
    "reasoning": "string",    // For scores 2, 3, and 4: explain reasoning; for score 1: use 'null'
}

### Scoring Criteria:
4 points: Key elements of the question and options are clearly visible, sufficient to directly answer the question.
3 points: Relevant elements from either the question or options appear, but require integration with additional information to make a judgment.
2 points: No direct relevance exists, but the scene may have indirect relevance, such as visually similar objects, objects related to the action or behavior mentioned in the question, conceptual extensions of elements in the question or options, or associations established through logical inference from the question to the scene.
1 point: Completely unrelated scene.

### Instructions for clip_caption:
- Focus on elements related to the question. Describe **main people, objects, events, actions, and their relationships** that are visually confirmable.
- If there are multiple scenarios, describe them respectively. Pay attention to the sequence of events!
- Only describe what is **directly observable**. Do **not** infer, imagine, or fabricate scenes beyond the visual evidence.
- If the question is about counting (e.g. 'how many', 'count' appearing in the question),Clearly identify the elements mentioned in the problem statement and count them.

### Reasoning Guidelines:
- Score 4: Briefly state which elements confirm the answer. Output the answer.
- Score 3: Explain what is missing or ambiguous (e.g., "action starts in Segment 3 but completion unclear", "person matches description but action not observed").
- Score 2: Explain how you decomposed or extended the question (e.g., "question asks about 'a musician', and a person holding a guitar appears").
- Score 1: Set reasoning to 'null'.

Be thorough, precise, and strictly grounded in visual evidence. Avoid temporal phrases like 'the first time'.
\end{lstlisting}
\end{tcolorbox}
\caption{Prompt for vlm\_scoring\_tool.}
\label{GA}
\end{figure*}